\documentclass[11pt,a4paper]{article}
\usepackage{times,latexsym}
\usepackage{url}
\usepackage[T1]{fontenc}

\usepackage[acceptedWithA]{tacl2018v2}
\usepackage[]{tacl2018v2}

\usepackage{xspace,mfirstuc,tabulary}

\newif\iftaclinstructions
\taclinstructionsfalse 
\iftaclinstructions
\renewcommand{\confidential}{}
\renewcommand{\anonsubtext}{(No author info supplied here, for consistency with
TACL-submission anonymization requirements)}
\newcommand{\instr}
\fi

\iftaclpubformat 

\else

\fi

\usepackage{bm}
\usepackage{here}
\usepackage[fleqn]{amsmath}
\usepackage{amssymb}
\usepackage{amsfonts}
\usepackage{multirow}
\usepackage{arydshln}
\usepackage{enumitem}
\usepackage{graphicx}
\usepackage{CJKutf8}
\usepackage{color}
\usepackage[utf8]{inputenc}
\usepackage{tikz}
\usepackage{multirow,makecell}

\def\tabref#1{Table~\ref{#1}}
\def\figref#1{Figure~\ref{#1}}
\def\equref#1{Equation~(\ref{#1})}
\def\secref#1{Section~\ref{#1}}

\newcommand\blfootnote[1]{%
  \begingroup
  \renewcommand\thefootnote{}\footnote{#1}%
  \addtocounter{footnote}{-1}%
  \endgroup
}

\def\revise#1{#1} 

\makeatletter
\newcommand{\pushright}[1]{\ifmeasuring@#1\else\omit$\displaystyle#1$\ignorespaces\fi}
\newcommand{\pushleft}[1]{\ifmeasuring@#1\else\omit$\displaystyle#1$\hfill\fi\ignorespaces}
\makeatother

\newcommand{\dashedline}[1]{%
    \tikz[baseline=(todotted.base)]{
        \node[inner sep=1pt,outer sep=0pt] (todotted) {#1};
        \draw[densely dashed] (todotted.south west) -- (todotted.south east);
    }%
}%

\def\MLE{MLE}
\def\RLDisREP{RL-D$_{{\rm REP}}$}
\def\RLDisDROP{RL-D$_{{\rm DROP}}$}

\def\RLGLEU{RL-GLEU}
\def\RLGLEUDisREP{RL-GLEU-D$_{{\rm REP}}$}
\def\RLGLEUDisDROP{RL-GLEU-D$_{{\rm DROP}}$}
\def\DisREP{D$_{{\rm REP}}$}
\def\DisDROP{D$_{{\rm DROP}}$}

\title{Neural Text Generation with Artificial Negative Examples}

\author{
  Keisuke Shirai$^\dagger$ ~~~ Kazuma Hashimoto$^{*\ddagger}$ ~~~ Akiko Eriguchi$^{*\mathsection}$ \\
  {\bf Takashi Ninomiya}$^\mathparagraph$ ~~~ {\bf Shinsuke Mori}$^\|$ \\
  $^\dagger$ Graduate School of Informatics, Kyoto University \\
  $^\ddagger$ Salesforce Research ~~~ $^\mathsection$ Microsoft Research \\
  $^\mathparagraph$ Graduate School of Science and Engineering, Ehime University \\
  $^\|$ Academic Center for Computing and Media Studies, Kyoto University \\
}

\date{}

\begin{document}
\maketitle

\begin{abstract}
Neural text generation models conditioning on given input (e.g. machine translation and image captioning) are usually trained by maximum likelihood estimation of target text. However, the trained models suffer from various types of errors at inference time. In this paper, we propose to suppress an arbitrary type of errors by training the text generation model in a reinforcement learning framework, where we use a trainable reward function that is capable of discriminating between references and sentences containing the targeted type of errors. We create such negative examples by artificially injecting the targeted errors to the references. In experiments, we focus on two error types, repeated and dropped tokens in model-generated text. The experimental results show that our method can suppress the generation errors and achieve significant improvements on two machine translation and two image captioning tasks.
\end{abstract}

\section{Introduction}~\label{sec-intro}
Conditional neural text generation models are expected to generate human-readable text that accurately describes source-side information~\citep{sutskever2014sequence,vinyals2015show}.~\blfootnote{$^*$Work done while the authors were at the University of Tokyo.} At training time, these models are trained in a supervised fashion by learning to predict ground-truth symbols~\citep{williams1989learning}, and at inference time, the models usually generate text in a left-to-right fashion. It is known that these text generation models suffer from a variety of generation errors. For example, the models repeat tokens unnecessarily and drop informative tokens; we call these two error types as {\it repeating error} and {\it dropping error}, respectively. Can we design a training framework to explicitly suppress a specific type of the generation errors?

We focus on a reinforcement learning (RL) framework proposed by \citet{ranzato2016sequence}, so that we can incorporate a reward function to penalize erroneous generation during training. The RL framework has been recently applied to many text generation tasks, and the authors have studied designing task-oriented reward functions~\citep{wu2016google,zhang2017sentence,rennie2017self}. More recent work~\citep{dai2017towards,gu2018neural} has proposed training a discriminative reward function, or simply a discriminator, in generative adversarial network frameworks. However, these reward functions are not designed to deal with a specific type of generation errors.

\begin{figure}[t]
  \centering
  \includegraphics[scale=0.5]{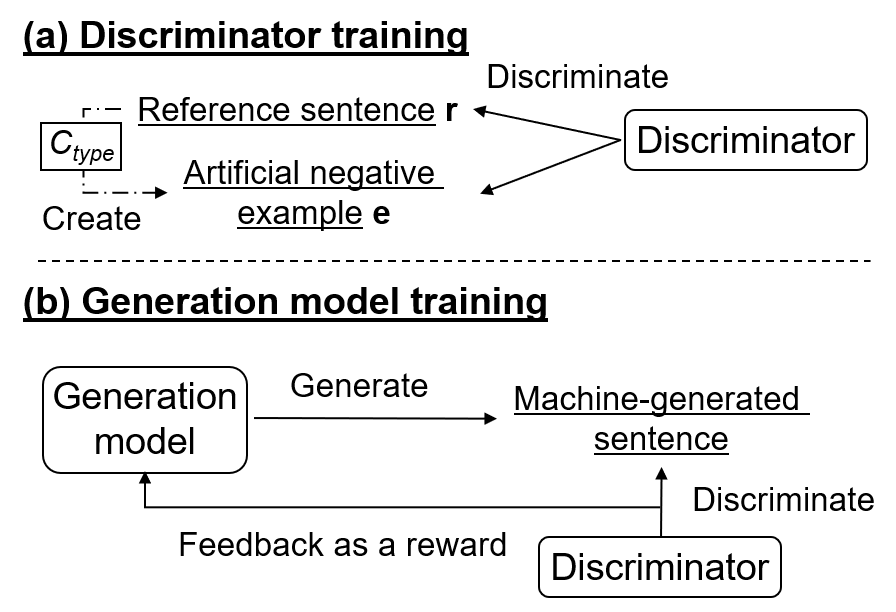} \\
  \caption{The overview of our method.}
  \label{fig-overview}
\end{figure}

To answer our research question, we propose a new RL framework that suppresses an arbitrary type of generation errors. \figref{fig-overview} illustrates the overview of our method. We first train a reward function that discriminates between references and sentences containing the targeted type of errors. To train such a discriminator, we introduce artificially-generated negative examples by injecting the specific type of errors to the references (\figref{fig-overview} (a)). The trained discriminator is expected to capture the targeted errors, and following \citet{ranzato2016sequence}, a text generation model learns to generate text while suppressing the errors (\figref{fig-overview} (b)). In this paper, we take two error types as examples: repeating and dropping errors. We show that our method can suppress these errors and improve generation performance on two translation and two captioning tasks.

Our contributions are three-fold:
\begin{itemize}
    \item We propose a novel RL framework that suppresses text generation errors by specifically targeting an error type. We show that our method can suppress two types of generation errors: the repeating and dropping errors.
    \item Our method can be expansively applied to the existent RL framework using other function such as GLEU, to further boost the generation performance. 
    \item Our analyses show that the proposed method works more effectively as training example size becomes smaller where more generation errors appear. 
\end{itemize}

\section{Neural text generation}\label{sec-generation}
\subsection{Maximum likelihood training}
In conditional neural text generation models, an encoder encodes source-side information $s$ (e.g. a source sentence in machine translation~\citep{sutskever2014sequence} and an image in image captioning~\citep{vinyals2015show}) into an intermediate representation $h^s$, and then a decoder is trained to generate a reference $r = (r_1, r_2, \cdots, r_n)$ conditioning on $h^s$. To train the model in a supervised fashion, we follow the maximum likelihood estimation (MLE) objective:
\begin{eqnarray}
  L_{\textrm{MLE}} = -\sum_{j=1}^{n} \log p (r_j | r_{< j}, s).
   \label{eq-mle}
\end{eqnarray} 
In this paper, we focus on two generation tasks: machine translation and image captioning.

At inference time, the text generation model generates a target-side sentence $t = (t_1, t_2, \cdots, t_{n'})$. When generating $j$-th token, the model follows a categorical distribution $p(t_j | t_{<j}, s)$ to generate a token. To generate a better sentence, we may use a variety of decoding techniques such as beam search.

\subsection{Reinforcement learning}\label{subsec-rl}
We next describe the RL framework proposed in \citet{ranzato2016sequence}. When training the text generation model in the RL framework, we consider the generation model as an agent, the categorical distribution at $j$-th decoding step $p(t_j | s, t_{<j})$ as a policy, where $t_j$ is the $j$-th token the model generates, and choosing a token from the categorical distribution as an action. After generating a complete target sentence $t = (t_1, t_2, \cdots, t_m)$, REINFORCE~\citep{williams1992simple} is used to define the loss function as follows:
\begin{equation}
\begin{split}
  L_{\textrm{RL}} = -\sum_{j=1}^{m} & \{ \log p(t_j | s, t_{<j}) \\
  & (R(s, t) - b (s, t_j)) \},
 \label{eq-rl}
\end{split}
\end{equation}
where $R$ is a reward function and $b$ is a baseline network~\citep{sutton2018reinforcement} to reduce the variance of the gradients.\footnote{Following \citet{ranzato2016sequence}, we simply use a linear regression model as the baseline network.} However, using only \equref{eq-rl} makes training unstable~\cite{wu2016google}. Thus, the following joint loss function is used:
\begin{eqnarray}
  \lambda_{\text{MIXED}} \ L_{\textrm{MLE}} + (1 - \lambda_{\text{MIXED}}) \ L_{\textrm{RL}},
  \label{eq-mixed}
\end{eqnarray}
where $\lambda_{\text{MIXED}}$ is a hyper-parameter to control the strength of the two signals.

\subsection{Generation errors}
Neural text generation models suffer from various types of errors such as repeating error, dropping error, grammatical error, misordering, and so on. Among them, the repeating and dropping errors are especially focused in the previous work~\citep{mi2016coverage, malaviya2018sparse,holtzman2020curious,welleck2020neural}. As a preliminary experiment, we trained a neural machine translation model with the MLE objective on the WAT'15 Workshop on Japanese-to-English translation task~\citep{nakazawa2016aspec}, and we manually checked $100$ examples that were randomly sampled from the translation results on the test dataset. We have found that $17$ sentences belong to the repeating error, where some tokens are repeated unnecessarily, and $13$ sentences belong to the dropping error, where some tokens drop from their reference sentences.\footnote{These $100$ translation examples are sampled from the RNMT result in \secref{subsec-results-dis}.} We also observed that the model trained with the widely-used RL approach in \citet{wu2016google} improves BLEU scores~\citep{papineni2002bleu} but does not necessarily suppress the above two types of errors.\footnote{Details will be shown in \secref{subsec-results-gleu}.} Based on these observations, we aim to improve the generation performance of the text generation model by suppressing two types of errors. 

\section{Approach}
We propose to suppress the targeted type of errors by using the RL framework. Our main focus is to design $R$ in \equref{eq-rl} that gives negative rewards to generated text if they contain the targeted type of errors, and positive rewards to the text otherwise. We use a discriminator to instantiate such an $R$ function. We train the discriminator by taking references as positive examples, and erroneous sentences as negative examples. We prepare these negative examples by artificially injecting the targeted type of errors to the references. In the remainder of this section, we describe the negative examples in \secref{subsec-ane} and the discriminator in \secref{subsec-dis}.

\subsection{Artificial negative examples}\label{subsec-ane}
We would like to prepare the discriminator that can specifically focus on one error type. The discriminator requires negative examples such that they always contain the targeted type of error. We create such negative examples directly from references by using an error-generating function $e = C_{type}(r)$, where $e$ is a sentence containing an error of a specific {\it type} and $r$ is a reference. We design $C_{type}$ depending on the error type we focus on. We call $e$ as an artificial negative example (ANE). 

In this study, we design two types of ANEs to deal with the repeating and dropping errors. \tabref{tab-examples-ane} shows a reference sentence and its negative examples of the two types.\footnote{While the tokens are split in a word-level in this example, we apply our method in a subword-level in our experiments.} In the following, we describe the design of ANEs.

\begin{description}[style=unboxed,leftmargin=0cm]

\item[Artificial repeating sentence]
We create artificial negative examples for the repeating errors by modifying reference sentences to repeat their tokens. In general, such repeated tokens do not always appear consecutively in model-generated sentences, and we propose the $C_{type}$ function as follows: given a reference $r$ whose length is $n$, $C_{repeat}(r)$ returns an artificial repeating sentence by duplicating $i$ consecutive tokens starting from the $j$-th token, at $k$ randomly-selected positions. We randomly choose $i$ from $(1, 2, \cdots, m_{\textrm{rep}})$, $j$ from $(1, 2, \cdots, n-i+1)$, and $k$ from $(1, 2, \cdots, n_{\textrm{rep}})$ for each example, where $m_{\textrm{rep}}$ is the maximum number of consecutive tokens, and $n_{\textrm{rep}}$ is the maximum number of the duplication. The $k$ random positions are selected from $(1, 2, \cdots, j-1, j+i-1, \cdots, n-1, n)$ not to break the original consecutive tokens. In \tabref{tab-examples-ane}, the three consecutive tokens ``by the company'' ($i=3$) starting from the $5$-th token ($j=5$) are repeated twice ($k=2$), and the tokens are inserted at after the $7$-th token, ``company,'' and the $9$-th token, ``keeps,'' of the reference. We set the hyper-parameters $(m_{\textrm{rep}},n_{\textrm{rep}}) = (4,4)$. We set $m_{\textrm{rep}} = n$ instead if $n$ is smaller than $n_{\textrm{rep}}$. 

\item[Artificial dropping sentence]
We create artificial negative examples for the dropping errors by modifying reference sentences to drop consecutive tokens. Given a reference $r$ whose length is $n$,  $C_{drop}(r)$ returns an artificial dropping sentence by dropping $i$ consecutive tokens starting from $j$-th token. We randomly choose $i$ from $(1, 2, \cdots, m_{\textrm{drop}})$ and $j$ from $(1, 2, \cdots, n-i+1)$ for each example, where $m_{\textrm{drop}}$ is the maximum number of consecutive tokens. In the example of \tabref{tab-examples-ane}, the three consecutive tokens ($i=3$) starting from the $5$-th token ($j=5$) are dropped. We set the hyper-parameter $m_{\textrm{drop}} = 4$. This setup allows us to drop randomly chosen ${i}$ consecutive tokens ($i = 1, 2, \cdots, 4$). We set $m_{\textrm{drop}} = n$ instead if $n$ is smaller than $m_{\textrm{drop}}$ and larger than $2$. $C_{drop}$ returns an end-of-sentence token if $n$ equals to $1$. 

\end{description}

\begin{table}[t]
  \footnotesize
  \begin{tabular}{cl}\hline
  
    \multirow{2}{*}{Reference} & A man is known \underline{by the company} \\
    & he keeps . \\[1.5mm]

    \multirowcell{3}{Artificial repeating \\ sentence} & A man is known \underline{by the company} \\
    & \underline{by the company} he keeps \underline{by the} \\
    & \underline{company} . \\[1.5mm]
    
    Artificial dropping & \multirow{2}{*}{A man is known he keeps . } \\
    sentence & \\\hline
    
  \end{tabular}
  \caption{Examples of artificial negative examples. Three consecutive tokens ``by the company'' are repeated in the repeating example, while they are dropped in the dropping example.}
  \label{tab-examples-ane}
\end{table}

\subsection{Discriminator}\label{subsec-dis}
We train the discriminator with references and their artificial negative examples. Our discriminator is a binary classifier that takes as input a pair of a source $s$ and its target sentence (either $r$ or $e = C_{type}(r)$) and outputs a real value in [0, 1]. We minimize the following loss function:
\begin{equation}
\begin{split}
  L_{\textrm{DIS}} 
    = &  -\mathbb{E}_{s, r}
    	  [\log \text{D}(s, r)]\\
    &  - \ \mathbb{E}_{s, e}
          [\log (1 - \text{D}(s, e))],
  \label{eq-dis}
  \end{split}
\end{equation}
where $\text{D}$ is a discriminator. During the discriminator training, we call $C_{type}(r)$ every time we process $r$ in a mini-batch. Once the discriminator training is finished, we use the trained discriminator $\text{D}$ in the RL framework by freezing its parameters, so that $R(s, t)$ can output a reward value for a generated sentence $t$ to train the text generation models.

Our discriminator $\text{D}$ consists of two types of encoder: a source-side and target-side encoders. The source-side encoder encodes a source-side information $s$ into a fixed-size vector $h^s$. The target-side encoder takes $h^s$ and encodes the target sentence $t = (t_1, t_2, \cdots, t_n)$ into a sequence of representations $H^t = (h^t_1, h^t_2, \cdots, h^t_n)$. Once $H^t$ is calculated, we compute $\hat{h}^t$ as $\hat{h}^t = maxpool(H^t)$. The discriminator finally obtains the output $y$ as:
\begin{equation*}
    y = f_{\text{sigmoid}}(W_o f_{\text{ReLU}}(W_h \hat{h}^t + b_h) + b_o),
\end{equation*}
where $W_h \in \mathbb{R}^{d_h/2 \times d_h}$, $b_h \in \mathbb{R}^{d_h/2}$, $W_o \in \mathbb{R}^{1 \times d_h/2}$, and ${b_o \in \mathbb{R}^1}$ are learnable parameters and $d_h$ is the dimension of $h^t$. $f_{\text{ReLU}}$ is the rectified linear unit (ReLU), and $f_{\text{sigmoid}}$ is the logistic sigmoid function.

\section{Experimental settings}

\subsection{Datasets}
In machine translation, we conducted experiments on Japanese-to-English (Ja-En) and German-to-English (De-En) tasks. For the Ja-En task, we used Asian scientific paper excerpt corpus (ASPEC) from the WAT'15 and used \texttt{train-1.txt} and \texttt{train-2.txt} for training, \texttt{dev.txt} for development, and \texttt{test.txt} for testing, following~\citet{hashimoto2019accelerated}. Both Japanese and English sentences were preprocessed as recommended in WAT'15.\footnote{\url{http://lotus.kuee.kyoto-u.ac.jp/WAT/WAT2015/baseline/dataPreparationJE.html}.} For the De-En task, we used the datasets provided by WMT'16\footnote{\url{http://www.statmt.org/wmt16/translation-task.html}.} and used all parallel corpora (\texttt{Europarl v7}, \texttt{Common Crawl corpus}, and \texttt{News Commentary v11}) for training, NewsTest2013 for development, and NewsTest2014 for testing.

In image captioning, we conducted experiments on two datasets: MS COCO~\citep{lin2014microsoft} and Flickr30K~\citep{plummer2015flickr30k}. For training, developing, and testing, we followed the splits provided by \citet{karpathy2015deep}.

We used {\tt SentencePiece}~\cite{kudo2018sentencepiece} for tokenizing sentences and building vocabularies. We empirically chose the vocabulary sizes of $16,000$ for each language in the Ja-En translation task, $16,000$ for both languages in the De-En translation task because these languages share an alphabet~\citep{sennrich2016neural}, and $2,000$ for MS COCO and Flickr30K captioning tasks. The vocabulary also contains three special tokens: <s> for beginning and </s> for end of a sentence, and <unk> for out-of-vocabulary tokens. \tabref{tab-datasets} shows the dataset statistics. During training in machine translation, we removed the sentence pairs whose maximum length is longer than $80$ and empty sentences from the training dataset. 

\begin{table}[t]
  \begin{center}
    \resizebox{.48\textwidth}{!}{
    \begin{tabular}{l|r|r|r|r|r}
      \multirow{2}{*}{dataset} & \multicolumn{2}{c|}{\vline \ \it{V} \vline \ } & 
        \multicolumn{3}{c}{\# examples} \\ \cline{2-6}
      & source & target & \multicolumn{1}{c|}{train} & \multicolumn{1}{c}{dev} & \multicolumn{1}{|c}{test} \\ \hline
      WAT'15 & 16,000 & 16,000 & 2,000,000 & 1,790 & 1,812 \\ 
      WMT'16 & \multicolumn{2}{c|}{16,000} & 4,548,880 & 3,000 & 3,003 \\ 
      MS COCO & \multicolumn{1}{|c|}{------} & 2,000 & 82,787 & 5,000 & 5,000 \\
      Flickr30K & \multicolumn{1}{|c|}{------} & 2,000 & 29,000 & 1,014 & 1,000 \\ \hline
    \end{tabular}
    }
    \caption{Dataset statistics and vocabulary sizes \vline \ \it{V} \vline.}
    \label{tab-datasets}
  \end{center}
\end{table}

\subsection{Models}
\subsubsection{Machine translation}
\paragraph{Text generation model}
For the translation tasks, we used two types of translation models: an RNN-based NMT (RNMT) model~\revise{\citep{bahdanau2015neural}} and the Transformer~\citep{vaswani2017attention}.
Our RNMT is an attention-based NMT model with a 2-layer bi-directional Long Short-Term Memory (LSTM)~\citep{hochreiter1997long} encoder and a 2-layer uni-directional LSTM decoder. 
\revise{We used the attention mechanism proposed by \citet{bahdanau2015neural}.}
The hidden state size and the embedding size were set to $512$.

Our Transformer consists of a $6$-layer encoder and a $6$-layer decoder. The hidden state size and the number of heads were set to $512$ and $8$, respectively. To add positional information, we used a positional encoding technique with sine and cosine functions proposed by \citet{vaswani2017attention}.

\paragraph{Discriminator}
For the discriminator, we used a 2-layer uni-directional LSTM for the source-side and target-side encoders with $d_h = 512$.

\subsubsection{Image captioning}
\paragraph{Text generation model}
For the image captioning tasks, we used a simple show-and-tell model~\citep{vinyals2015show}. We used ResNet-152~\citep{he2016deep} pre-trained on an ImageNet classification dataset~\citep{russakovsky2015imagenet} as an encoder by freezing its model parameters to only extract features, and a $512$-dimensional $1$-layer uni-directional LSTM as a decoder. Once an image feature $f^i \in \mathbb{R}^{2048}$ is extracted, the decoder state $h^t_0$ is initialized as $h^t_0 = W_i f^i + b_i$, where $W_i \in \mathbb{R}^{512 \times 2048}$ is a weight matrix and $b_i \in \mathbb{R}^{512}$ is a bias vector.

\paragraph{Discriminator}
For the discriminator, we project the feature vector $f^i$ into the fixed-size vector $h^s$ as $h^s = W_D f^i + b_D$, where a weight matrix $W_D \in \mathbb{R}^{512 \times 2048}$ and a bias vector $b_D \in \mathbb{R}^{512}$ are learnable parameters. We used a uni-directional LSTM as the target-side encoder with $d_h = 512$.

\subsection{Training strategies}
For all the experiments, we used a single GPU of \texttt{NVIDIA GeForce GTX 1080 Ti}.

\subsubsection{Training discriminators}
We used Adam~\citep{kingma2015adam} with the initial learning rate of $1.0 \times 10^{-3}$ and weight decay with the rate of $1.0 \times 10^{-6}$. We checked its accuracy on the development dataset every $1,000$ iterations, halved the learning rate when the accuracy went worse, and finished the training when the learning rate was halved for five times. We chose the best models based on the best accuracy on the development dataset. For the development dataset, we created one negative example for each reference sentence. 

\subsubsection{Training text generation models}
The training of the text generation models can be divided into two steps: a pre-training step with the MLE loss \equref{eq-mle} and a reinforcement learning step with the RL loss \equref{eq-mixed} with the pre-trained model following \citet{wu2016google}.

In the pre-training step, we used Adam with the initial learning rate of $1.0 \times 10^{-3}$ and weight decay with the rate of $1.0 \times 10^{-6}$ unless stated otherwise. Each mini-batch contains $128$ examples. We checked its perplexity on the development dataset every $1,000$ iterations and halved the learning rate if the perplexity went worse. We finished the training when we halved the learning rate for five times. For the Transformer, we used AdamW~\citep{loshchilov2019decoupled} with weight decay with the rate of $1.0 \times 10^{-4}$. When using AdamW, we scheduled the learning rate as follows:
\begin{align}
  \pushleft{lr =} & \nonumber\\
  \begin{cases}
    lr_{\text{ini}} + step \times \frac{lr_{\text{max}} - lr_{\text{ini}}}{N_{\text{wm}}} & \text{if}\ step \leq N_{\text{wm}} \\
    lr_{\text{max}} \times \eta(step) & \text{otherwise} 
  \end{cases},
  \label{eq:cosine}
\end{align}
where
\begin{equation}
    \eta(step) = 0.5 + 0.5 \times \text{cos}(\pi \times \frac{step - N_{\text{wm}}}{N_\text{wm} \times N_{\text{cl}}}). \nonumber
\end{equation}
This scheduling consists of two steps: a linear warmup step from $lr_{\text{ini}}$ to $lr_{\text{max}}$ for the first $N_{\text{wm}}$ iterations, and a cosine annealing step from $lr_{\text{max}}$ to $0$ for $N_{\text{wm}} \times N_{\text{cs}}$ iterations. 
We used empirically tuned hyper-parameters as $(lr_{\text{ini}}, lr_{\text{max}}, N_{\text{wm}}, N_{\text{cs}}) = (5.0 \times 10^{-6}, 5.0 \times 10^{-4}, 4,000, 24)$. Each mini-batch contains $512$ examples for the Transformer.
In the pre-training step, we clipped the gradients~\citep{pascanu2013difficulty} with a value of $1.0$ and used the label-smoothing technique~\citep{szegedy2016rethinking} with the rate of $0.1$. We chose the best models based on the lowest perplexity on the development dataset.

In the reinforcement learning step, we used stochastic gradient decent with momentum. \revise{For the translation tasks, we tuned a learning rate of per model with a momentum rate of $0.9$. For the captioning tasks, we consistently used a learning rate of $5.0 \times 10^{-2}$ with a momentum rate of $0.9$. In this step, we took $20,000$ iterations for fine-tuning the parameters of the generation model.} We continued to use the same gradient clipping and label-smoothing techniques. Each mini-batch contains $64$ examples. In particular, we carefully tuned $\lambda_{\text{mixed}}$ in \equref{eq-mixed}.\footnote{More concretely, we searched the values in a coarse-to-fine fashion in \{0.5, 0.3, 0.1, $7.5 \times 10^{-2}$, $5.0 \times 10^{-2}$, $2.5 \times 10^{-2}$, $1.0 \times 10^{-2}$, $7.5 \times 10^{-3}$, $5.0 \times 10^{-3}$, $2.5 \times 10^{-3}$, $1.0 \times 10^{-3}$\}.} We report our model inference generated by the beam search with the width of $10$.

\subsection{Evaluation metrics}
One goal in this study is to suppress the repeating and dropping errors. To quantitatively evaluate these errors in the model's output, we used REP and DROP scores~\citep{malaviya2018sparse}. In this section, we first describe these two metrics in detail, then introduce task-specific metrics.

\subsubsection{REP score}
The REP score calculates how many $n$-gram repetitions are included in a model-generated sentence $t$, given its reference $r$, as follows:
\begin{equation*}
  \text{REP}(r, t) = \frac{\sigma(t,r)}{ \sum_{w \in V} r(ww) + \sum_{s \in V^{n}_r} r(s)},
\end{equation*}
where
\begin{eqnarray}
  \sigma(t,r) = \lambda_2 \underset{\hspace{0.2cm} s \in V^{n}_r, t(s) \geq 2}{\sum} \hspace{-0.25cm} \max \{0, t(s) - r(s)\} \nonumber \\
  + \lambda_1 \sum_{w \in V} \max \{0, t(ww) - r(ww)\}. 
  \label{eq-rep}
\end{eqnarray}
$V^{n}_r$ is the set of all the $n$-grams included in the reference. $r(ww)$ and $r(s)$ show the frequency of consecutive $1$-gram $w$ and $n$-gram $s$ of the reference, respectively, and $t(ww)$ and $t(s)$ show those of the machine-generated sentence $t$. $\lambda_1$ and $\lambda_2$ are hyper-parameters.

The REP score is defined for each $n$-gram separately, but in this study, we propose to use an extended REP (eREP) score that evaluates consecutive $1$-gram and $n$-grams ($n=2,3,4$) together. 
The eREP score is calculated as follows:
\begin{equation*}
  \text{eREP}(r, t) = \frac{\sigma(t,r)}{ \sum_{w \in V} t(ww) + \sum_{s \in V^{n}_t} t(s)},
\end{equation*}
where
\begin{eqnarray*}
  \sigma(t,r) = \sum^{4}_{n=2} \lambda_n\underset{\hspace{0.2cm} s \in V^{n}_t, t(s) \geq 2}{\sum} \hspace{-0.25cm} \max \{0, t(s) - r(s)\} \nonumber \\
  + \lambda_1 \sum_{w \in V} \max \{0, t(ww) - r(ww)\}.
\end{eqnarray*}
$V^{n}_t$ is the set of all the $n$-grams included in the machine-generated sentence $t$. 
We weight the repetitions of $n$-grams equally ($\lambda_n=1$). 
For the eREP score, lower is better.

\subsubsection{DROP score}
In the machine translation tasks, the DROP score calculates how many source tokens are {\it not} covered in model-generated sentences. A word alignment tool is used to identify which source tokens are aligned with its reference, and calculates the ratio of those not aligned with the generated text (hypothesis). In other words, this aims at evaluating how well the source information is covered, and the score is defined as follows:
\begin{equation*}
  \text{DROP}(c_{\text{ref}}, c_{\text{hyp}}) = 1 - \frac{1}{|c_{\text{ref}}|} \sum_{i \in c_{\text{ref}}} in(i),
\end{equation*}
where $c_{\text{ref}}$ and $c_{\text{hyp}}$ represent the set of source tokens' indices in the source-reference and source-hypothesis alignments, respectively. $in(i)$ is a function that returns $1$ if an index $i$ is included in $c_{\text{hyp}}$ and $0$ otherwise. For the DROP score, lower is better similarly as in the eREP score.

In the captioning tasks, however, the DROP score is not applicable because the source sentence does not exist. In the experiment, we used ROUGE$_L$~\citep{chen2015microsoft} instead to evaluate how many dropping errors are found in the model's output. In ROUGE$_L$, higher is better contrary to the DROP score. In our preliminary experiment, to check the relationship between the DROP score and ROUGE$_L$, we calculated the Pearson correlation coefficient between the ROUGE$_L$ and the DROP on the RNMT's WAT'15 Ja-En result. We confirmed a negative moderate correlation ($-0.430$) between these two metrics, which means that we can use ROUGE$_L$ instead of the DROP score in the captioning tasks. 

\begin{table*}[t]
  \begin{center}
    \footnotesize
    \begin{tabular}{l|ll|cccc}
            \multicolumn{1}{c|}{Task} &
            \multicolumn{2}{c|}{Model} &
            eREP ($\downarrow$) &
            DROP ($\downarrow$) & 
            BLEU (BP) & 
            METEOR \\\hline
            
        \multirow{6}{*}{WAT'15 Ja-En} & 
            \multirow{3}{*}{\underline{RNMT}} &

              	\MLE & 
                	\revise{2.78}~~~ & 
                	\revise{15.03}~~~ & 
                    \revise{25.28~~~ (1.000)} & 
                    \revise{31.31}~~~ \\\cdashline{4-7}[1pt/2pt]
                & & \RLDisREP & 
                	\revise{{\bf 2.53}~$\dagger$} & 
                	\revise{15.03}~~~ & 
                    \revise{{\bf 25.50}~~~ (1.000)} & 
                    \revise{31.19}~~~ \\
                & & \RLDisDROP & 
                	\revise{2.67}~~~ & 
                	\revise{{\bf 14.63}}~$\dagger$ & 
                    \revise{25.24~~~ (1.000)} & 
                    \revise{{\bf 31.34}}~~~ \\\cline{2-7}
                    
            & \multirow{3}{*}{\underline{Transformer}} &
              	\MLE & 
                	2.00~~~ & 
                	12.46~~~ & 
                    \revise{28.58~~~ (1.000)} &
                    \revise{33.09}~~~ \\\cdashline{4-7}[1pt/2pt]
                & & \RLDisREP & 
                	{\bf 1.83}~\revise{$\dagger$} & 
                	12.24~\revise{$\dagger$} & 
                    \revise{{\bf 28.93~$\dagger$} (1.000)} &
                    \revise{33.27~$\dagger$} \\
                & & \RLDisDROP & 
                	2.03~~~ & 
                	{\bf 11.95}~\revise{$\dagger$} & 
                    \revise{28.75~~~ (1.000)} &
                    \revise{{\bf 33.45}~$\dagger$} \\\hline
        \multirow{6}{*}{WMT'16 De-En} & 
            \multirow{3}{*}{\underline{RNMT}} &

            	\MLE & 
            	    \revise{1.09}~~~ &
            	    ~ \revise{3.87}~~~ &
            	    \revise{{\bf 24.65}~~~ (1.000)} &
                    \revise{29.71}~~~ \\\cdashline{4-7}[1pt/2pt]
                & & \RLDisREP & 
            	    \revise{{\bf 0.78}}~~~ &
            	    ~ \revise{4.07}~~~ &
            	    \revise{24.60~~~ (0.993)} &
            	    \revise{29.57}~~~ \\
                & & \RLDisDROP & 
            	    \revise{0.89}~~~ &
            	    ~ \revise{{\bf 3.59}~$\dagger$} &
            	    \revise{24.56~~~ (1.000)} &
                    \revise{{\bf 29.87}}~$\dagger$ \\\cline{2-7}
                    
            & \multirow{3}{*}{\underline{Transformer}} &
              	\MLE & 
                	{\bf 0.30}~~~ & 
                	~ 3.45~~~ & 
                    \revise{27.19~~~ (0.973)} &
                    \revise{31.64}~~~ \\\cdashline{4-7}[1pt/2pt]
                & & \RLDisREP & 
                	0.31~~~ & 
                	~ 3.36~\revise{$\dagger$} & 
                    \revise{{\bf 27.39}~$\dagger$ (0.973)} &
                    \revise{31.72~$\dagger$} \\
                & & \RLDisDROP & 
                	0.31~~~ & 
                	~ {\bf 3.17}~\revise{$\dagger$} & 
                    \revise{27.35~~~ (0.979)} &
                    \revise{{\bf 31.78}~$\dagger$} \\\hline
    \end{tabular}
  \caption{Results on the translation tasks.}
  \label{tab-results-translation}
  \end{center}
\end{table*} 
\begin{table*}[t]
  \begin{center}
    \footnotesize
    \begin{tabular}{l|l|ccccccc}
            \multicolumn{1}{c|}{Task} &
            \multicolumn{1}{c|}{Model} &
            eREP ($\downarrow$) &
            ROUGE$_L$ ($\uparrow$) & 
            BLEU-1 & 
            BLEU-2 & 
            BLEU-3 & 
            BLEU-4 & 
            CIDEr \\\hline
        
        \multirow{3}{*}{MS COCO} & 
          	\MLE & 
                \revise{7.42}~~~ & 
                \revise{49.80}~~~ & 
                \revise{64.71}~~~ & 
                \revise{47.09}~~~ & 
                \revise{34.41}~~~ & 
                \revise{25.69}~~~ & 
                \revise{82.77}~~~ \\\cdashline{3-9}[1pt/2pt]
            & \RLDisREP & 
                \revise{{\bf 6.51}~$\dagger$} & 
                \revise{{\bf 49.95}}~~~ & 
                \revise{{\bf 66.16}~$\dagger$} & 
                \revise{{\bf 48.21}~$\dagger$} & 
                \revise{{\bf 35.35}~$\dagger$} & 
                \revise{{\bf 26.54}~$\dagger$} & 
                \revise{{\bf 83.06}}~~~ \\
            & \RLDisDROP & 
                \revise{7.59}~~~ & 
                \revise{49.63}~~~ & 
                \revise{64.20}~~~ & 
                \revise{46.36}~~~ & 
                \revise{33.69}~~~ & 
                \revise{25.02}~~~ & 
                \revise{82.02}~~~ \\\hline
                
        \multirow{3}{*}{Flickr30K} & 
            \MLE & 
                \revise{9.25}~~~ & 
                \revise{43.35}~~~ & 
                \revise{60.52}~~~ & 
                \revise{41.78}~~~ & 
                \revise{28.64}~~~ & 
                \revise{19.47}~~~ & 
                \revise{41.24}~~~ \\\cdashline{3-9}[1pt/2pt]
            & \RLDisREP & 
                \revise{{\bf 5.51}~$\dagger$} & 
                \revise{42.53}~~~ & 
                \revise{{\bf 62.90}~$\dagger$} & 
                \revise{{\bf 43.80}~$\dagger$} & 
                \revise{{\bf 30.29}~$\dagger$} & 
                \revise{{\bf 20.76}}~~~ & 
                \revise{40.75}~~~ \\
            & \RLDisDROP & 
                \revise{6.80~$\dagger$} & 
                \revise{{\bf 43.54}}~~~ & 
                \revise{61.74}~~~ & 
                \revise{43.20}~~~ & 
                \revise{29.93}~~~ & 
                \revise{20.63}~~~ & 
                \revise{{\bf 42.29}}~~~ \\\hline
    \end{tabular}
  \caption{Results on the captioning tasks.}
  \label{tab-results-captioning}
  \end{center}
\end{table*}
\begin{table}[t]
  \begin{center}
    \footnotesize
    \begin{tabular}{l|cc}
            \multicolumn{1}{c|}{Task} &
            \DisREP &
            \DisDROP \\\hline
        
        WAT'15 Ja-En & 96.34 & 77.60 \\ 
        WMT'16 De-En & 96.77 & 73.08 \\ 
        MS COCO      & 99.14 & 87.25 \\ 
        Flickr30K    & 98.63 & 82.17 \\\hline
    \end{tabular}
  \caption{Accuracy of the discriminator on the development dataset.}
  \label{tab-results-discriminator}
  \end{center}
\end{table}
\begin{table*}[t]
  \footnotesize
  \begin{tabular}{c|l}\hline
  
    \multicolumn{2}{l}{Example (A)} \\ \hline
    \multirow{2}{*}{Source} & \begin{CJK}{UTF8}{ipxm} 
    Ｉの主なものにシメチジン，ラニチジン，ファモチジンなどがあり，１日１回投与と２回 \end{CJK} \\
    & \begin{CJK}{UTF8}{ipxm} 投与で治癒率に差を認めない。\end{CJK} \\[1.5mm]
    
    \multirow{2}{*}{Reference} & There are Cimetidine, Ranitidine, Famotidine, etc. in I, and differences are not recognized \\
    & at therapeutic ratio in first 1 time administration and 2 time administration. \\[1.5mm] 
    
    \multirow{2}{*}{\MLE} & There were cimetidine, ranitidine, famotidine, etc. in the main thing of I, \underline{and the difference was not} \\
    & \underline{recognized} at 1 time administration and 2 time administration, \underline{and the difference was not recognized}. \\[1.5mm]
    
    \multirow{2}{*}{\RLDisREP} & There were cimetidine, ranitidine, famotidine, etc. in the main thing of I, and the difference was not \\
    & recognized at 1 time administration and 2 time administration. \\[1.5mm] \hline\hline
    
    \multicolumn{2}{l}{Example (B)} \\ \hline
    
    \multirow{2}{*}{Source} & \multirow{2}{*}{\begin{CJK}{UTF8}{ipxm} その結果，\dashedline{電気光学特性として，}Ｖ１０＝１１．８Ｖ，Ｖ９０＝１８Ｖを得た \end{CJK}} \\[1.5mm]
    
    \multirow{2}{*}{Reference} & \multirow{2}{*}{\dashedline{The electro‐optical property} obtained was V10=11.8V and V90=18V.} \\[1.5mm]
    
    \multirow{2}{*}{\MLE} & \multirow{2}{*}{As a result, V10=11.8V and V90=18V were obtained.}\\[1.5mm]
    
    \multirow{2}{*}{\RLDisDROP} & \multirow{2}{*}{As a result, V10=11.8V and V90=18V were obtained \dashedline{as the electro‐optic characteristics.}} \\\\\hline
    
  \end{tabular}
  \caption{Generation examples of the RNMT-based models on the Ja-En translation task. The underlined tokens with solid lines represent the repeating error. The underlined tokens with a dashed line are tokens \MLE\ failed to generate.}
  \label{tab-examples-wat15}
\end{table*}

\subsubsection{Task-specific metrics}
We used BLEU and METEOR~\citep{malaviya2018sparse} for the translation tasks, and BLEU and CIDEr~\citep{vedantam2015cider} for the captioning tasks. For the captioning tasks, we used the publicly available code\footnote{\url{https://github.com/tylin/coco-caption}.} to calculate the BLEU and CIDEr scores. Note that for the captioning tasks, we report BLEU-{1,2,3,4}. 

\subsection{Model configurations}
\begin{description}[style=unboxed,leftmargin=0cm]
\item[-- \MLE] is the baseline model trained by the MLE loss in \equref{eq-mle}.

\item[-- \RLDisREP, \RLDisDROP] are our proposed models trained by the RL framework in \equref{eq-mixed} with our proposed discriminator. 
The text generation model parameters are initialized by the \MLE baseline. \RLDisREP\ and \RLDisDROP\ are trained to suppress the repeating and dropping errors with the discriminator \DisREP\ and \DisDROP, respectively. 

\end{description}

\section{Results}\label{sec-results}
\revise{
We report the scores in [0, 100]. The symbol~$\dagger$ follows a score if a system produces a significant improvement against \MLE\ (p < $0.05$). In our experiments, a statistical significance test was performed by the paired bootstrap resampling method~\citep{koehn2004statistical}.
}

\subsection{Can we suppress the targeted errors?}\label{subsec-results-dis}
In this section, we show the results of using the discriminator to suppress the repeating and dropping errors. For the \RLDisREP\ and \RLDisDROP\ models, we chose the best models based on the score on the development datasets we would like to improve (e.g. the eREP score used for \RLDisREP). 

We first show the accuracy of our discriminators because the discriminator plays the key role in our method. \tabref{tab-results-discriminator} reports the binary classification accuracy on the development datasets. \DisREP\ achieves over $96$\% accuracy in all the tasks, while the accuracy of \DisDROP\ is not as high as that of \DisREP. The possible reason for this is that identifying the dropped tokens in a sentence is a more difficult task than identifying the $n$-gram repetitions.

\tabref{tab-results-translation} shows the main results on the translation tasks. \RLDisREP\ consistently improves the eREP scores, except that the MLE-based Transformer in the De-En task has much less room for improvement. \RLDisDROP\ also consistently improves the DROP scores in both the RNMT and Transformer models. Receiving reward from the discriminator, namely \DisREP\ and \DisDROP, the text generation models learn to suppress the repeating and dropping errors\revise{, and this answers the research question we argued in \secref{sec-intro}}. We can also see that \RLDisREP\ tends to increase the BLEU and that \RLDisDROP\ tends to increase the METEOR \revise{though some exceptions exist}. For the former, we consider that reducing the amount of repetitions by \RLDisREP\ leads to higher $n$-gram precision and thus the model can achieve higher BLEU scores. For the latter, this is presumably because the METEOR is based on unigram precision and recall, and it puts more weight on recall than on precision. Therefore, suppressing the dropping errors and restoring dropped tokens by \RLDisDROP\ contribute to the improvements of the METEOR.

\tabref{tab-results-captioning} shows the results on the image captioning tasks. Again, we can see that \RLDisREP\ significantly improves the eREP scores, leading to the better BLEU-{1,2,3,4} scores in both the COCO and Flickr30K tasks. On the other hand, \RLDisDROP\ does not achieve significant improvements on the ROUGE$_L$. One possible reason is due to the diversity of text in captioning task~\citep{dai2017towards}. In other words, identifying whether the dropping error occurs in a sentence would be an easy task for the discriminator, but generating such tokens would be difficult for the generation model. 
We also observed that the balancing factor $\lambda_{\text{mixed}}$ is preferred to be smaller for \RLDisREP\ than that for \RLDisDROP. This indicates that the generation model can rely more on the reward from \DisREP\ than that from \DisDROP, which supports why \RLDisREP\ produces more expected results than \RLDisDROP\ in the captioning tasks.\footnote{For example, we took $5.0 \times 10^{-3}$ and $5.0 \times 10^{-3}$ for \RLDisREP, and $0.1$ and $5.0 \times 10^{-2}$ for \RLDisDROP\ on the COCO and Flickr30K tasks, respectively.} 

We show generation examples of the RNMT-based models in the WAT'15 Ja-En translation task in \tabref{tab-examples-ane}. In Example (A), \RLDisREP\ successfully suppresses the repeated tokens that are found in \MLE\ and generates the non-repetitive sentence. In Example (B), \RLDisDROP\ satisfactorily restores the dropped tokens in \MLE\ and generates more informative sentence.

\begin{table*}[t]
  \begin{center}
    \footnotesize
    \begin{tabular}{l|ll|cccc}
            \multicolumn{1}{c|}{Task} &
            \multicolumn{2}{c|}{Model} &
            eREP ($\downarrow$) &
            DROP ($\downarrow$) & 
            BLEU (BP) & 
            METEOR \\\hline
            
        \multirow{8}{*}{WAT'15 Ja-En} & 
            \multirow{4}{*}{\underline{RNMT}} &

                \MLE & 
                    \revise{2.78}~~~ &
                    \revise{15.03}~~~ &
                    \revise{25.28~~~ (1.000)} &
                    \revise{31.31}~~~ \\
                & & \RLGLEU & 
                    \revise{2.84}~~~ &
                    \revise{13.37~$\dagger$} &
                    \revise{25.73~~~ (1.000)} &
                    \revise{{\bf 32.19}~$\dagger$} \\\cdashline{4-7}[1pt/2pt]
                & & \RLGLEUDisREP & 
                    \revise{{\bf 2.65}}~~~ &
                    \revise{13.81~$\dagger$} &
                    \revise{{\bf 25.83}~$\dagger$ (1.000)} &
                    \revise{31.94~$\dagger$} \\
                & & \RLGLEUDisDROP & 
                    \revise{2.96}~~~ &
                    \revise{{\bf 13.13}~$\dagger$} &
                    \revise{25.76~$\dagger$ (1.000)} &
                    \revise{32.28~$\dagger$} \\\cline{2-7}
                
            & \multirow{4}{*}{\underline{Transformer}} &
                \MLE & 
                	2.00~~~ & 
                	12.46~~~  & 
                    \revise{28.58~~~  (1.000)} &
                    \revise{33.09}~~~  \\
                & & \RLGLEU & 
                	2.02~~~  & 
                	11.11~\revise{$\dagger$} & 
                    \revise{28.67~~~ (1.000)} & 
                    \revise{33.77~$\dagger$} \\
                    \cdashline{4-7}[1pt/2pt]
                & & \RLGLEUDisREP & 
                	{\bf 1.94}~~~ & 
                	11.30~\revise{$\dagger$} & 
                    \revise{{\bf 28.99}~~~ (1.000)} &
                    \revise{33.75~$\dagger$} \\
                & & \RLGLEUDisDROP & 
                	1.96~~~ & 
                	{\bf 11.05}~\revise{$\dagger$} & 
                    \revise{28.92~~~ (1.000)} &
                    \revise{{\bf 33.86}~$\dagger$} \\\hline
                    
        \multirow{8}{*}{WMT'16 De-En} & 
            \multirow{4}{*}{\underline{RNMT}} &

                \MLE & 
                    \revise{1.09}~~~ &
                    ~ \revise{3.87}~~~ &
                    \revise{24.65~~~ (1.000)} &
                    \revise{29.71}~~~ \\
                & & \RLGLEU & 
                    \revise{0.65~$\dagger$} &
                    ~ \revise{{\bf 2.76}~$\dagger$} &
                    \revise{24.57~~~ (1.000)} &
                    \revise{30.08~$\dagger$} \\\cdashline{4-7}[1pt/2pt]
                & & \RLGLEUDisREP & 
                    \revise{{\bf 0.59}~$\dagger$} &
                    ~ \revise{2.86~$\dagger$} &
                    \revise{24.70~~~ (0.998)} &
                    \revise{30.01~$\dagger$} \\
                & & \RLGLEUDisDROP & 
                    \revise{0.68~~~} &
                    ~ \revise{2.85~$\dagger$} &
                    \revise{{\bf 24.72}~~~ (1.000)} &
                    \revise{{\bf 30.11}~$\dagger$} \\\cline{2-7}
                
            & \multirow{4}{*}{\underline{Transformer}} &
                \MLE & 
                	0.30~~~ & 
                	~ 3.45~~~ & 
                	\revise{27.19~~~ (0.973)} &
                	\revise{31.64}~~~ \\
                & & \RLGLEU & 
                	{\bf 0.29}~~~ & 
                	~ {\bf 3.08}~\revise{$\dagger$} & 
                	\revise{27.13~~~ (0.973)} &
                	\revise{{\bf 31.74}~$\dagger$} \\\cdashline{4-7}[1pt/2pt]
                & & \RLGLEUDisREP & 
                	0.30~~~ & 
                	~ 3.20~\revise{$\dagger$} & 
                	\revise{{\bf 27.20}~~~ (0.970)} &
                	\revise{31.69~$\dagger$} \\
                & & \RLGLEUDisDROP & 
                	{\bf 0.29}~~~ & 
                	~ 3.10~~~ & 
                	\revise{27.09~~~ (0.974)} &
                	\revise{31.69~$\dagger$} \\\hline
    \end{tabular}
  \caption{Results of the joint models with GLEU on the translation tasks. Note that the \MLE\ results are the same in \tabref{tab-results-translation}.}
  \label{tab-results-translation-gleu}
  \end{center}
\end{table*}
\begin{table*}[t]
  \begin{center}
    \footnotesize
    \begin{tabular}{l|l|ccccccc}
            \multicolumn{1}{c|}{Task} &
            \multicolumn{1}{c|}{Model} &
            eREP ($\downarrow$) &
            ROUGE$_L$ ($\uparrow$) & 
            BLEU-1 & 
            BLEU-2 & 
            BLEU-3 & 
            BLEU-4 & 
            CIDEr \\\hline
        
        \multirow{4}{*}{MS COCO} & 
            \MLE & 
                ~ \revise{{\bf 7.42}}~~~ & 
                \revise{49.80}~~~ & 
                \revise{64.71}~~~ & 
                \revise{47.09}~~~ & 
                \revise{34.41}~~~ & 
                \revise{25.69}~~~ & 
                \revise{82.77}~~~ \\
            & \RLGLEU & 
                \revise{10.65}~~~ & 
                \revise{51.99~$\dagger$} & 
                \revise{68.83~$\dagger$} & 
                \revise{52.05~$\dagger$} & 
                \revise{38.28~$\dagger$} & 
                \revise{28.06~$\dagger$} & 
                \revise{87.87~$\dagger$} \\\cdashline{3-9}[1pt/2pt]
            & \RLGLEUDisREP & 
                ~ \revise{7.96}~~~ & 
                \revise{{\bf 52.09}~$\dagger$} & 
                \revise{{\bf 70.31}~$\dagger$} & 
                \revise{{\bf 53.09}~$\dagger$} & 
                \revise{{\bf 39.01}~$\dagger$} & 
                \revise{{\bf 28.74}~$\dagger$} & 
                \revise{89.11~$\dagger$} \\
            & \RLGLEUDisDROP & 
                ~ \revise{9.04}~~~ & 
                \revise{51.99~$\dagger$} & 
                \revise{69.06~$\dagger$} & 
                \revise{51.97~$\dagger$} & 
                \revise{38.21~$\dagger$} & 
                \revise{28.21~$\dagger$} & 
                \revise{{\bf 89.41}~$\dagger$} \\\hline
                
        \multirow{4}{*}{Flickr30K} & 
            \MLE & 
                ~ \revise{9.25}~~~ & 
                \revise{43.35}~~~ & 
                \revise{60.52}~~~ & 
                \revise{41.78}~~~ & 
                \revise{28.64}~~~ & 
                \revise{19.47}~~~ & 
                \revise{41.24}~~~ \\
            & \RLGLEU & 
                ~ \revise{5.65~$\dagger$} & 
                \revise{{\bf 44.08}~$\dagger$} & 
                \revise{63.02~$\dagger$} & 
                \revise{44.41~$\dagger$} & 
                \revise{30.79~$\dagger$} & 
                \revise{21.19~$\dagger$} & 
                \revise{{\bf 41.54}}~~~ \\\cdashline{3-9}[1pt/2pt]
            & \RLGLEUDisREP & 
                ~ \revise{5.32~$\dagger$} & 
                \revise{43.97}~~~ & 
                \revise{{\bf 63.62}~$\dagger$} & 
                \revise{{\bf 44.86}~$\dagger$} & 
                \revise{{\bf 31.12}~$\dagger$} & 
                \revise{{\bf 21.34}~$\dagger$} & 
                \revise{41.27}~~~ \\
            & \RLGLEUDisDROP & 
                ~ \revise{{\bf 5.05}~$\dagger$} & 
                \revise{43.82}~~~ & 
                \revise{63.42~$\dagger$} & 
                \revise{44.50~$\dagger$} & 
                \revise{30.67~$\dagger$} & 
                \revise{20.95~$\dagger$} & 
                \revise{41.10}~~~ \\\hline
    \end{tabular}
  \caption{Results of the joint models with GLEU on the captioning tasks. Note that the \MLE\ results are the same in \tabref{tab-results-captioning}.}
  \label{tab-results-captioning-gleu}
  \end{center}
\end{table*}

\begin{figure*}[t]

  \begin{minipage}{0.5\linewidth}
    \centering
    \includegraphics[scale=0.5]{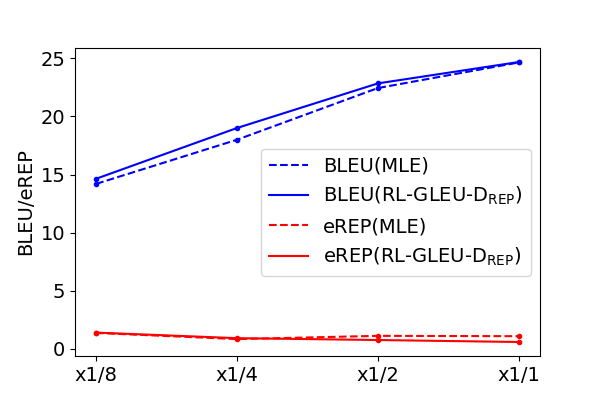} \\
    \ \ \ \ (a) eREP and BLEU (RNMT)
  \end{minipage}
  \begin{minipage}{0.5\linewidth}
    \centering
    \includegraphics[scale=0.5]{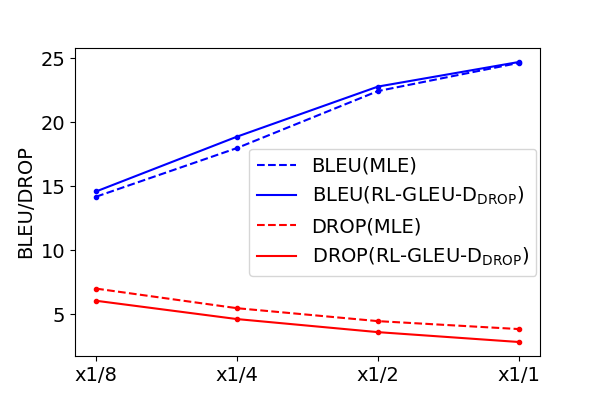} \\
    \ \ \ \ (b) DROP and BLEU (RNMT)
  \end{minipage}
  \begin{minipage}{0.5\linewidth}
    \centering
    \includegraphics[scale=0.5]{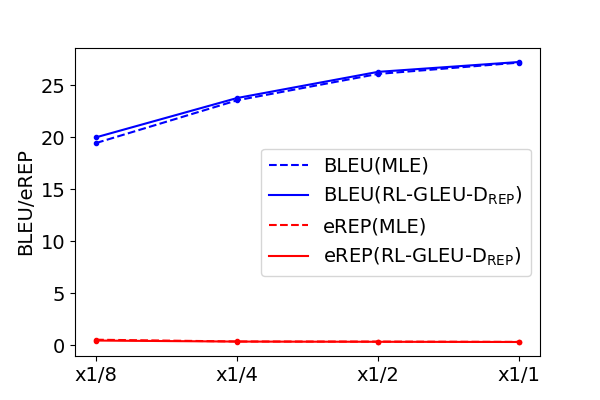} \\
    \ \ \ \ (c) eREP and BLEU (Transformer)
  \end{minipage}
  \begin{minipage}{0.5\linewidth}
    \centering
    \includegraphics[scale=0.5]{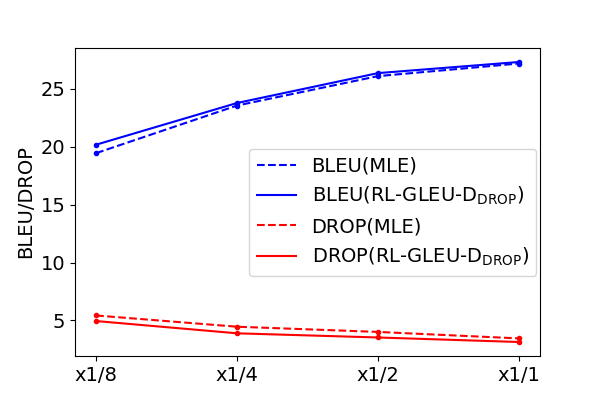} \\
    \ \ \ \ (d) DROP and BLEU (Transformer)
  \end{minipage}
  
  \caption{The results on the WMT'16 De-En translation task when varying the training size exponentially.}
  \label{fig-scale}
  
\end{figure*}

\subsection{Can we incorporate an off-the-shelf reward function?}\label{subsec-results-gleu}
In \secref{subsec-results-dis}, we have shown that our proposed discriminator can suppress its targeted type of errors. We investigate how existing reward functions work compared with our discriminators, to further improve the generation performance by incorporating them. Taking GLEU~\citep{wu2016google} as an example of existent rewards, we propose the following joint reward function:
\begin{eqnarray}
    R(s, t) & = & \lambda_{\text{RL}} R'(s, t) \nonumber \\
            &   & + (1 - \lambda_{\text{RL}}) GLEU(t, r), 
    \label{eq-reward}
\end{eqnarray}
where $R'$ is one of our reward functions and $GLEU$ is the GLEU score. The GLEU is known to be effective in improving the BLEU~\citep{wu2016google}. $GLEU(t, r)$ calculates the minimum of the generated sentence $t$'s $n$-gram precision and recall against the reference $r$. $\lambda_{\text{RL}}$ is a hyper-parameter to control the strength of the two signals. In this section, we refer \RLGLEU\ to the model that uses only the GLEU as the reward, \RLGLEUDisREP\ (or \RLGLEUDisDROP) to the model that uses both the GLEU and the corresponding discriminator. We chose the best models based on the best BLEU score of the development datasets.


\tabref{tab-results-translation-gleu} shows the results on the translation tasks. In the Ja-En task, \RLGLEU\ contributes to the improvement on the BLEU, DROP, and METEOR scores. Since the GLEU computes recall against the reference, \RLGLEU\ consequently generated more (informative) tokens and improved those scores. This is why \RLGLEUDisDROP\ less improves the DROP this time. \RLGLEUDisREP\ further improves the eREP, especially when \RLGLEU\ deteriorates the eREP score. In the De-En task, the Transformer-based \RLGLEUDisREP\ and \RLGLEUDisDROP\ show almost the same results of \RLGLEU. This might be due to a large training examples, and we will discuss the effect of training data size in \secref{subsec-results-scale}.


\tabref{tab-results-captioning-gleu} shows the results on the image captioning tasks. In both the COCO and Flickr30K tasks, \RLGLEU\ greatly improves the BLEU and the CIDEr, but generates more repetitions as the eREP indicates. \RLGLEUDisREP\ achieves further improvements on the BLEU scores by suppressing the repeating errors. \RLGLEUDisDROP, on the other hand, does not boost the performance of \RLGLEU. We consider that it is difficult to suppress the dropping errors even when using both the GLEU and the discriminator's signals as we discussed in \secref{subsec-results-dis}. 

\begin{table*}[t]
  \begin{center}
    \footnotesize
    \begin{tabular}{l|l|cccc}
            \multicolumn{1}{c|}{Task} &
            Model &
            eREP ($\downarrow$) &
            DROP ($\downarrow$) & 
            BLEU (BP) & 
            METEOR \\\hline
            
        \multirow{3}{*}{WAT'15 Ja-En} & 
            
              	\MLE & 
                	\revise{2.78}~~~ & 
                	\revise{15.03}~~~ & 
                    \revise{25.28~~~ (1.000)} & 
                    \revise{31.31}~~~ \\\cdashline{3-6}[1pt/2pt]
                & CovVec & 
                    \revise{2.51}~~~ &
                    \revise{14.97}~~~ &
                    \revise{25.22~~~ (1.000)} &
                    \revise{31.25}~~~ \\
                & GGD & 
                    \revise{2.88}~~~ &
                    \revise{14.98}~~~ &
                    \revise{25.41~~~ (1.000)} &
                    \revise{31.48~$\dagger$} \\\hline
                    
        \multirow{3}{*}{WMT'16 De-En} & 
            
            	\MLE & 
            	    \revise{1.09}~~~ &
            	    \revise{3.87}~~~ &
            	    \revise{24.65~~~ (1.000)} &
                    \revise{29.71}~~~ \\\cdashline{3-6}[1pt/2pt]
                & CovVec & 
                    \revise{0.86}~~~ &
                    \revise{3.79}~~~ &
                    \revise{24.66~~~ (1.000)} &
                    \revise{29.58}~~~ \\
                & GGD & 
                    \revise{0.87}~~~ &
                    \revise{3.74~$\dagger$} &
                    \revise{24.75~~~ (1.000)} &
                    \revise{29.79~$\dagger$} \\\hline

    \end{tabular}
  \caption{Results of the coverage-based and GAN-based models. Note that the \MLE\ results are the same in \tabref{tab-results-translation}.}
  \label{tab-results-translation-additional}
  \end{center}
\end{table*}
\revise{
\subsection{Comparison with related studies}
In the field of machine translation, there are two types of studies that share similarities with our research direction: (i) the method based on coverage and (ii) the method based on Generative Adversarial Network (GAN). The coverage-based method~\citep{mi2016coverage,tu2017context,malaviya2018sparse} that uses attention history to reduce the amount of the repeating and dropping errors. The GAN-based method~\citep{gu2018neural,yang2018improving} utilizes an adversarial training with a discriminator for training the generation model to generate more natural sentences. In this section, we conduct further experiments to compare these methods with ours. Throughout this section, we focus on the RNMT model.
}

\revise{
\subsubsection{Comparison with the coverage-based method}
We first conduct experiments of the coverage-based model. We decided to use the coverage vector~\citep{tu2017context} that stores the attention history and is used in the decoding process to more effectively utilize untranslated source words. We used the Neural Network based coverage, which uses RNNs to model the coverage vector, and set $10$ to the coverage dimension following \citep{tu2017context}. Although \citet{tu2017context} uses a Gated Recurrent Unit (GRU) as an activation function, we used LSTM and empirically confirmed that LSTM also works well. We refer to this model as CovVec. 
}

\revise{
\tabref{tab-results-translation-additional} shows the results. In both the two translation tasks, we can see that CovVec successfully improves the eREP and DROP scores from the \MLE\ results as we expected. In comparison with \tabref{tab-results-translation}, the eREP scores are close to those of \RLDisREP\ ($2.53$ in the Ja-En task and $0.78$ in the De-En task), and the drop scores are reasonably higher than those of \RLDisDROP\ ($14.63$ in the Ja-En task and $3.59$ in the De-En task), respectively. However, CovVec does not produce any significant improvements on the BLEU and METEOR scores in our experiments, though \RLDisREP\ in the Ja-En task achieved the improvement on BLEU. We consider that this improvement comes from the fact that, by fine-tuning the parameters of \MLE, our model can only concentrate on suppressing the target type of errors for erroneous sentences and keep generating the same sentences for other non-problematic sentences. When we checked the translations of \RLDisREP\ and \MLE, we found that \RLDisREP\ tends to generat the almost same translations of \MLE\ for some source sentences supporting our view.
}

\revise{
\subsubsection{Comparison with the GAN-based method}
We next conduct experiments of the GAN-based model. We decided to use the Gumbel-Greedy Decoding (GGD)~\citep{gu2018neural} that bridges the generation model and the discriminator by using the Gumbel-Softmax estimator~\citep{jang2017categorical}. The hyper-parameters ($N_g$, $N_d$) in \citep{gu2018neural} were set to ($1$, $1$). We refer to this model as GGD. 
}

\revise{
The results are shown in \tabref{tab-results-translation-additional}. For the BLEU and METEOR, GGD consistently improves the both scores from the \MLE\ results. For the eREP and DROP scores, GGD harms the eREP and improve the DROP in the Ja-En task, while GGD improves the both scores in the De-En task. This indicates that the generation model of GGD does not always learn to suppress both repeating and dropping errors. One possible reason for this is that a type of errors the discriminator deals with can differ, as the frequency of each error type included in machine-generated sentences changes depending on the task and the generation model.
}

\subsection{Varying training size}\label{subsec-results-scale}
We have confirmed that our proposed method works effectively to suppress the targeted type of errors on the WAT'15 Ja-En translation and the COCO and the Flickr30K captioning tasks, but is less effective in the WMT'16 De-En task especially during the joint training with the GLEU. One presumable reason is that the training dataset size in the De-En translation task would be large enough to train the language model well and to suppress the targeted errors that are caused by data sparseness. We hypothesize that our method would be more effective especially when the training dataset is small; in such a situation, the language model would not be trained well and would produce more errors. 

To verify this hypothesis, we conducted experiments by varying the training dataset size exponentially (1/1, 1/2, 1/4, 1/8) on the WMT'16 De-En translation task. For simplicity, we changed only the training dataset size, and set the same hyper-parameters used in \secref{subsec-results-gleu}. \figref{fig-scale} shows the results. Note that the 1/1 case is the same as that in \secref{subsec-results-gleu}. \figref{fig-scale} (a), (b), (d) show that the results meet our expectation; the smaller the training dataset size is, the more effective our strategy is in the eREP, DROP, and BLEU scores. In \figref{fig-scale} (c), by contrast, \RLGLEUDisREP\ is less effective because the Transformer language model might be strong enough to suppress most of the repetitions even with the smallest training dataset.


\section{Related work}\label{sec-results-discussion}
Training neural text generation models with a discriminator have recently been studied in generative adversarial network based methods~\citep{dai2017towards,gu2018neural,yang2018unsupervised}. 
\revise{
These work utilize the generative adversarial framework~\citep{goodfellow2014generative,arjovsky2017wasserstein} to train the sentence generation model for generating more natural, human-like sentences.
}
Although our framework can also be considered as such a generator-discriminator framework, there are two points that differentiate ours from the GAN-based methods. One is that our generator can focus on suppressing a targeted type of errors as the discriminator learns to discriminate them from references with artificially-generated erroneous sentences. The other is that our artificial negative examples definitely contain errors, while the machine-generated sentences in the GAN-based methods can be correct sentences as they are sampled from the generation model.

\revise{
There are several studies that use negative examples for training a sentence generation model. To focus on the repetitions, \citet{welleck2020neural} proposed the unlikelihood training with the training objective that decreases the probability of generating the token which has appeared in the previous context. 
\citet{he2020negative} collected the negative examples from the model generations and trained the model not to generate them. 
One advantage of ours is that our discriminator is trained independent of the sentence generation model as our method does not require any extra step of collecting negative examples from the sentence generation model.
}

Coverage-based methods in machine translation has also aimed at suppressing the repeating and dropping errors by focusing on coverage information about source words~\citep{mi2016coverage,tu2017context,malaviya2018sparse}. 
\revise{
While these methods deal with the same types of errors that we focused on the experiments, our method is not limited to these error types and we can apply our method to other error types by designing the artificial negative examples.
}

Automatic post-editing aims at editing machine-translated sentences to fix errors, writing styles, and so on~\citep{freitag2019ape,gu2019levenshtein}. The task is similar to ours, in that they deal with how to make the model-generated sentences better. However, there are two significant differences. One is that they use a machine translation model and a post-editing model independently, while we consistently train one model to generate less erroneous sentences. The other is that, our model explicitly focuses on one specific generation error type of interest, which is not feasible in the previous work.



\section{Conclusion}
We have proposed a reinforcement learning based method that suppresses a targeted type of errors. Our method uses a discriminator that captures a specific type of errors and the discriminator is trained with artificially-generated negative examples. In experimental results, we have shown that our method can suppress repeating and dropping errors.
In future work, it is interesting to explore application of our method to source-side unannotated data.

\bibliography{tacl2018}
\bibliographystyle{acl_natbib}

\end{document}